\documentclass[manuscript,screen]{acmart}
\usepackage{xcolor}
\usepackage{amsmath}
\usepackage{subcaption}
\usepackage{graphicx}
\usepackage{multirow}
\usepackage{makecell}

\AtBeginDocument{%
  \providecommand\BibTeX{{%
    \normalfont B\kern-0.5em{\scshape i\kern-0.25em b}\kern-0.8em\TeX}}}

\setcopyright{acmlicensed}
\copyrightyear{2024}
\acmYear{2024}
\acmDOI{}

\acmConference[HT '24]{34th ACM Conference on Hypertext and Social Media}{September 10--13, 2024}{Poznan, Poland}
\acmISBN{978-1-4503-XXXX-X/18/06}

\begin{document}

\title[Seeing Through AI's Lens]{Seeing Through AI's Lens: Enhancing Human Skepticism Towards LLM-Generated Fake News}

\author{Navid Ayoobi}
\author{Sadat Shahriar}
\author{Arjun Mukherjee}
\affiliation{%
  \institution{University of Houston}
  \city{Houston}
  \state{Texas}
  \country{USA}
  \postcode{43017-6221}
}


\begin{abstract}
LLMs offer valuable capabilities, yet they can be utilized by malicious users to disseminate deceptive information and generate fake news. The growing prevalence of LLMs poses difficulties in crafting detection approaches that remain effective across various text domains. Additionally, the absence of precautionary measures for AI-generated news on online social platforms is concerning. Therefore, there is an urgent need to improve people's ability to differentiate between news articles written by humans and those produced by LLMs. By providing cues in human-written and LLM-generated news, we can help individuals increase their skepticism towards fake LLM-generated news. This paper aims to elucidate simple markers that help individuals distinguish between articles penned by humans and those created by LLMs. To achieve this, we initially collected a dataset comprising 39k news articles authored by humans or generated by four distinct LLMs with varying degrees of fake. We then devise a metric named Entropy-Shift Authorship Signature (ESAS) based on the information theory and entropy principles. The proposed ESAS ranks terms or entities, like POS tagging, within news articles based on their relevance in discerning article authorship. We demonstrate the effectiveness of our metric by showing the high accuracy attained by a basic method, i.e., TF-IDF combined with logistic regression classifier, using a small set of terms with the highest ESAS score. Consequently, we introduce and scrutinize these top ESAS-ranked terms to aid individuals in strengthening their skepticism towards LLM-generated fake news.
\end{abstract}




\maketitle
\section{Introduction}
The proliferation of large language models (LLMs) has facilitated the generation of text that closely resembles human writing in terms of quality.
As a result, LLMs are utilized across various domains like chatbots \cite{zheng2024judging}, code generation \cite{liu2024your,amoozadeh2023trust}, translation \cite{yao2023empowering}, and text summarization \cite{pu2023summarization}.
Despite the beneficial roles they play, LLMs present opportunities for misuse by malicious users, enabling the dissemination of misinformation or deception.
Consequently, this misuse can exacerbate the prevalence of fake reviews, disruptions in educational systems, and the erosion of trust among users of online social networks \cite{ayoobi2023looming,radivojevic2024llms}.
Furthermore, the generative and reasoning capabilities of LLMs render them effective tools for the generation of fake news, typically with the intent of promoting specific political agendas, influence electoral outcomes, and undermining democratic principles.

The pursuit of developing precise detection methods emerges as an essential solution in combating the misuse of LLMs and alleviating their consequential impacts.
Currently, there is increasing research dedicated to exploring techniques for improving the accuracy of such detection systems.
Literature indicates that in scenarios where there is consistency in the topic of the articles present in the training data and the testing data, and where the same LLM generates the text for both training and testing articles, detection methods exhibit notably high levels of accuracy.
This phenomenon primarily stems from the decoding strategies inherent to LLMs, along with the statistical patterns they impart to detection systems \cite{pagnoni2022threat}.
Conversely, human evaluators often exhibit diminished accuracy (about $70\%$) in discerning AI-generated text \cite{ippolito2019automatic} compared to detectors trained on extensive datasets.
This discrepancy arises from humans' inability to track common and shared patterns among all generated sample texts.
Consequently, the virtually indistinguishable nature of generated text at the surface level presents significant challenges for human identification.
On the other hand, altering the text domain or employing multiple LLMs for text generation can deviate the distribution that the LLM relies on for predicting subsequent tokens, thus potentially misleading detectors trained on specific text domains and resulting in a substantial decline in their performance \cite{bhattacharjee2024eagle}.
These unreliable detectors may unintentionally deceive individuals, hindering their skepticism towards fake news.

Watermarking was introduced as a method to enable the detection of any text generated by an LLM, irrespective of its domain \cite{kirchenbauer2023watermark,kuditipudi2023robust}.
Watermarking involves embedding a concealed pattern within the syntactic text, facilitating its algorithmic identification.
The drawback associated with watermarking resides in its susceptibility to removal through paraphrasing.
Moreover, if the watermarking procedure is open-sourced, imposters retain the ability to disrupt the watermark following the generation of fabricated content.
On the other hand, if the watermarking strategy remains closed-source, it becomes imperceptible to regular users, with only model developers possessing awareness of its existence.
Hence, this situation remains unchanged for individuals exposed to fake news on the platforms other than the one where the watermarking strategy is known.

The rise of abundant number of LLMs from multiple research groups \cite{zhao2023survey} makes it further challenging to devise a detection approach that accommodates the majority of these LLMs.
Further aggravating the situation is the absence of automatic detection systems in online social networks to flag potential LLM-generated articles.
Hence, there is an urgent need to enhance individuals' proficiency in distinguishing between articles penned by humans and those generated by LLMs.
By offering some hints in human-authored news and news generated by LLMs, we can assist people in fostering a more critical perspective on the source of synthetic fake news.

In this paper, we focus on elucidating straightforward indicators that enable individuals to discern whether an article is authored by a human or an LLM.
Our methodology comprises four key stages. 
Initially, we present our collected dataset, comprising 3000 news articles sourced from trustworthy outlets and encompassing diverse topics such as sports, celebrities, history and religion, politics and government, social culture and civil rights, science and information technology. 
Utilizing four LLMs, we produce fake versions of the genuine news at three levels of fake, resulting in a dataset of 39k news articles. 
To the best of our knowledge, this is the first large dataset that can be used for analyzing fake news generated by different LLMs at varying degrees of fake.
We make this dataset publicly available to facilitate further research in this area.
In the second stage, we devise a metric termed as Entropy-Shift Authorship Signature (ESAS), leveraging information theory and entropy principles.
This metric ranks the terms or other entities, like Part-of-Speech (POS) tagging, within the news article based on their significance in identifying the authorship of an article as either human-written or LLM-generated.
In the third stage, we demonstrate the efficacy of the proposed ESAS metric in identifying significant cues within news articles. 
This is evidenced by the high accuracy achieved by a basic approach, specifically TF-IDF \cite{ramos2003using} combined with logistic regression classifier, when fed with a small set of terms with the highest ESAS score.
The resulting accuracy exceeds current human capabilities in detecting AI-generated content.
We believe that incorporating these top ESAS-ranked terms can heighten human skepticism towards LLM-generated fake news.
Consequently, in the fourth stage, we delve into and highlight the most crucial terms under diverse conditions, taking into account various LLMs and fake levels.

The primary contributions of this paper can be summarized as follows:
\begin{itemize}
    \item  We collect a reasonably large dataset comprising 39k news articles. These articles are either authored by humans or generated by four different LLMs, exhibiting varying degrees of fake news.
    \item We introduce the ESAS metric, which ranks terms or entities, like POS tagging, within news articles based on their relevance to identifying article authorship.
    \item We offer news readers cues to identify the authorship of articles, enhancing their ability to detect fake news crafted by LLMs.
\end{itemize}

\section{Related work}
Numerous studies have concentrated on identifying fake news by examining linguistic patterns \cite{rashkin2017truth,seddari2022hybrid,choudhary2021linguistic}, fact-checking methodologies \cite{pan2018content,hu2021compare}, understanding writer intentions \cite{giachanou2022impact}, and sentiment analysis \cite{cui2019same,alonso2021sentiment}. 
Choudhary \textit{et al.} \cite{choudhary2021linguistic} designed a neural network model for fake news detection, utilizing syntactic, grammatical, sentimental, and readability features from news articles. Their findings revealed that a model trained exclusively on syntax, sentiment, and grammar-based features outperformed one trained on readability-based features. 
Authors in \cite{seddari2022hybrid} proposed a set of knowledge-based features, introduced as \textit{fact-verification} features, which include the credibility of the news website, the count of sources publishing the news, and the views of well-known fact-checking websites. 
By incorporating these with linguistic features, they achieved a $94\%$ accuracy rate on the Buzzfeed Political News dataset.
\textit{SAME}, an end-to-end deep learning framework, was proposed by Cui \textit{et al.} \cite{cui2019same} to assess the latent sentiments present in user comments on social media, potentially aiding in the discrimination of fake news from reliable information.
Giachanou \textit{et al.} \cite{giachanou2022impact} approached the issue from a psycholinguistic perspective.
The authors utilized the Convolutional Neural Network (CNN) and introduced the \textit{CheckerOrSpreader} model.
This model aims to distinguish between users who propagate fake news (spreaders) and those who challenge it (checkers).
Their research outcomes revealed that checkers generally use more positive language and a richer vocabulary, whereas spreaders use more informal language, such as slang and swear words.

The emergence of language models has opened up new avenues for imposters to produce vast amounts of fake news across various platforms, attempting to sway public opinion and persuade them with their LLM-generated misleading content \cite{pagnoni2022threat,zellers2019defending,sun2024exploring,hu2024bad,kareem2023fighting,jiang2024disinformation,wu2023fake,su2023fake}.
Recognizing the limitations of current detection systems can assist in identifying the risks posed by different imposters using LLMs.
Pagnoni \textit{et al.} \cite{pagnoni2022threat} explored three threat scenarios based on an adversary's budget and expertise, affecting the text generation style.
These scenarios include text generated using available LM APIs, utilizing a pre-trained model with potential parameter modifications, and an LM model fine-tuned on specific data.
The effectiveness of different detection systems was evaluated across these scenarios.
Wu and Hooi \cite{wu2023fake} proposed \textit{SheepDog} as a solution to counteract the deterioration of existing detection methods after LLM Style Attacks, where imposters mimic the style of trusted news sources using LLMs.
SheepDog is a style-independent fake news detector that leverages LLMs to produce different reframings for each news article, introducing a range of styles.
Subsequently, it maximizes the consistency in the model's predictions across the original article and its diverse reframings.
In contrast to conventional detection methods that rely on machine learning and deep learning, Jiang \textit{et al.} \cite{jiang2024disinformation} explored the potential of LLMs to detect the fake content they produce. 
The findings indicated that ChatGPT struggles to effectively detect the disinformation it generates.
However, by crafting a tailored prompt, its detection accuracy can be moderately improved.

While the emphasis of most studies has been on enhancing detector accuracy, the absence of warning systems for AI-generated news on online platforms remains a reality. 
With the rapid introduction of more LLMs over time, this could not have been more timely to educate people about potential indicators in news articles to heighten their skepticism towards fake LLM-generated news.
Therefore, in this paper, we propose simple cues that individuals can employ to distinguish between human-authored and LLM-generated news.

\section{Dataset}
For the purpose of this study, a total of 200k news articles sourced from reputable outlets, including ABC News, Aljazeera, American Press, Associated Press News, CBS News, CNN, NBC News, Reuters, and The Guardian, were gathered.
The BERTopic \cite{grootendorst2022bertopic} model was employed to filter and select $3000$ articles, ensuring their primary focus aligned with one of the following five topics: sports, celebrities, history and religion, politics and government, social culture and civil rights, science and information technology.
To ensure dataset balance, each of the aforementioned topics is represented by an equal sample size of 500 articles.

Four models, specifically GPT 3.5, Mistral-7B \cite{jiang2023mistral}, Llama2-7B, and Llama2-13B \cite{touvron2023Llama} were employed to generate LLM counterparts of the original articles.
To obtain different level of fake, three strategies were devised for generating LLM samples: \textbf{1. Article Rephrasing, 2. Article Extending, and 3. Summary Expanding}.
The first strategy, article rephrasing, involved prompting the LLM with \textit{``Assume you are a journalist in a news agency. Rephrase following news article''}, and subsequently inputting the original article.
The second strategy, article extending, entailed determining the total word count of the original article and using the prompt \textit{``Assume you are a journalist in a news agency. Extend and complete following news article until it reaches a length of [N] words''}.
The placeholder [N] was substituted with the total word count of the original article, and the first third of the original article was then fed as input to be expanded by the LLM.
The third strategy, summary expanding, involved a combination of summary and expansion. 
Initially, the prompt \textit{``Summarize following news article''} was used to condense the original news article.
Subsequently, the total word count of the original article was calculated.
The resulting summary, along with the prompt \textit{``Assume you are a journalist in a news agency. Write a news article that comprises [N] words based on the following summarized news article''}, was used to guide the LM in generating a final sample.
To maintain length consistency, the placeholder [N] was substituted with the total word count of the original article prior to summarization.
The dataset collected will be made available to the research community, enabling researchers to pursue further investigations on this field\footnote{\url{https://github.com/navid-aub/News-Dataset}}.

\section{Methodology}
In this section, we employ principles derived from information theory \cite{aizawa2003information}, specifically utilizing mutual information, to derive a metric of term significance in news articles.
This metric serves to quantify the level of uncertainty between the terms utilized within a news article and its attributed authorship, distinguishing between human-authored news and that produced by an LLM.
Through the establishment of this metric, we ascertain the systematic ranking of terms predicated upon their discriminatory efficacy.
Such systematic ranking underscores the pivotal terms that aid individuals in cultivating skepticism regarding the provenance of news articles they read, particularly in discerning whether they originate from an LLM or human sources.
Initially, we demonstrate that a straightforward approach, namely term frequency-inverse document frequency (TF-IDF) \cite{ramos2003using} coupled with a basic classifier, like logistic regression, exhibits a remarkable capability to discern LLM-generated news articles when it utilizes only a limited number of terms ranked based on the proposed metric.
Therefore, by introducing and analyzing these terms, we anticipate a reduction in human susceptibility to falling into the trap of fake news propagated by LLMs.
By familiarizing themselves with the characteristic terms present in LLM-generated news articles, individuals can enhance their discernment and accuracy in distinguishing between authentic and fabricated news.

\subsection{Term awareness and uncertainty reduction}
The quantification of information pertaining to an event with probability $P(\mathcal{X}{=}x)$ is denoted by $\log(\frac{1}{P(\mathcal{X}=x)})$.
Self-entropy, $H(\mathcal{X})$, representing the extent of uncertainty pertaining to a random variable $\mathcal{X}$, is determined by the expected amount of information associated to $\mathcal{X}$,
\begin{equation}
\begin{split}
    H(\mathcal{X}) = \sum_{x \in \mathcal{X}}P(\mathcal{X}{=}x)\log(\frac{1}{P(\mathcal{X}{=}x)})
    \end{split}
\end{equation}
Let $\mathcal{A}$ be a random variable indicating the categorization of news article authorship, wherein two discrete outcomes are delineated: $a_1$ representing human-authored news, and $a_2$, denoting LLM authorship.
Let $\mathcal{W}$ denote a random variable representing the constituent words within a news article.
The permissible outcomes of $\mathcal{W}$ encompass the set of distinct lexical entities, $w_i$'s, drawn from the aggregate vocabulary derived from both human-written news articles and those generated by LLMs.
The pairwise mutual information, $M(a_i,w_j)$,concerning authorship $a_i$ and word $w_j$ is expressed as the discrepancy in information content between the joint probability $P(\mathcal{A}{=}a_i,\mathcal{W}{=}w_j)$ and the product of the probabilities $P(\mathcal{A}{=}a_i)$ and $P(\mathcal{W}{=}w_j)$ under the assumption of independence between authorship $a_i$ and word $w_j$,
\begin{equation}
\begin{split}
    M(a_i,w_j) = \log \frac{P(\mathcal{A}{=}a_i,\mathcal{W}{=}w_j)}{P(\mathcal{A}{=}a_i)P(\mathcal{W}{=}w_j)}
    \end{split}
\end{equation}
The expected mutual information, denoted by $I(\mathcal{A};\mathcal{W})$, serves as a metric quantifying the reduction in uncertainty with respect to authorship $\mathcal{A}$ consequent to the awareness of the words comprising the news article,
\begin{equation}\label{I}
\begin{split}
    &I(\mathcal{A};\mathcal{W}) = \sum_{w \in \mathcal{W}} \sum_{a \in \mathcal{A}} P(a,w)\log(\frac{P(a,w)}{P(a)P(w)})\\
    &= \sum_{w \in \mathcal{W}} \sum_{a \in \mathcal{A}} P(a)P(w|a)\log(\frac{1}{P(a)})-P(w)P(a|w)\log(\frac{1}{P(a|w)}) \\
    &= \sum_{a \in \mathcal{A}} P(a)\log(\frac{1}{P(y)}) - \sum_{w \in \mathcal{W}} P(w)\sum_{a \in \mathcal{A}} P(a|w)\log(\frac{1}{P(a|w)}) \\ 
    &= H(\mathcal{A})-H(\mathcal{A}|\mathcal{W})
\end{split}
\end{equation}
Equation \ref{I} validates that $I(\mathcal{A};\mathcal{W})$ encapsulates the difference in entropy pertaining to authorship before and after focusing on attributes exclusively provided by the words contained within a news article.
Consequently, the capacity of each word within the vocabulary to serve as an indicator of the authorship of a news article can be quantified and ranked accordingly.
To achieve this objective, we represent the mutual information as the summation over all words,
\begin{equation}
\begin{split}
    I(\mathcal{A};\mathcal{W}) &= \sum_{w_i \in \mathcal{W}} P(w_i)\Bigl(H(\mathcal{A})-H(\mathcal{A}|\mathcal{W}=w_i)\Bigl) \\
    &\stackrel{\text{def}}{=} \sum_{w_i \in \mathcal{W}} I_i(\mathcal{A};\mathcal{W}=w_i)
\end{split}
\end{equation}
We term $I_i$ as the Entropy-Shift Authorship Signature (ESAS).
Utilizing the ESAS metric enables the prioritization of words in the vocabulary of news articles, highlighting the most crucial ones for human readers in distinguishing between human-written news and news generated by LLMs.

\subsection{Estimating probabilities in ESAS metric equations}
Let $\mathcal{D}{=}\mathcal{D}_L \cup \mathcal{D}_H$ denote the collection of all news articles composed by both LLMs and humans, where $\mathcal{D}_L$ represents news articles generated by LLMs and $\mathcal{D}_H$ is the set of news articles authored by humans.
The total word count within $\mathcal{D}$ is assumed to be N.
The set of distinct words within $\mathcal{D}$ is represented as $\mathcal{W}$, and the frequency of the $i^{th}$ word in $\mathcal{W}$, denoted as $w_i$, within $\mathcal{D}$ is computed as $N_i{=}N_{L,i}{+}N_{H,i}$, where $N_{L,i}$ and $N_{H,i}$ denote the frequency of $w_i$ in news articles generated by LLMs and those written by humans, respectively.
The probability $P(w_i)$ represents the likelihood of occurrence of $w_i$ in a news article and can thus be approximated as the ratio $\frac{N_i}{N}$.
To compute $H(\mathcal{A})$, we assume a uniform probability distribution over the random variable $\mathcal{A}$. Consequently, a news article is considered either human or LLM-generated with equal probability, $P(\mathcal{A}{=}a_1)=P(\mathcal{A}{=}a_2){=}\frac{1}{2}$, and $H(\mathcal{A})$ is computed as $\log2$.
In order to calculate the conditional entropy $H(\mathcal{A}|\mathcal{W}{=}w_i)$, it is necessary to compute the conditional probability $P(a_j|w_i)$. 
This probability is determined by dividing the number of occurrences of $w_i$ in authorship category $a_j$ ($j=1,2$) by the total occurrences of $w_i$ in both authorship categories (LLM-generated and human-authored news articles).
\begin{equation}
\begin{split}
    H(\mathcal{A}|\mathcal{W}{=}w_i) &{=} -P(a_1|w_i)\log(P(a_1|w_i)){-}P(a_2|w_i)\log(P(a_2|w_i)) \\
    &{=} \frac{N_{L,i}}{N_i} \log(\frac{N_i}{N_{L,i}})+\frac{N_{H,i}}{N_i} \log(\frac{N_i}{N_{H,i}})
\end{split}
\end{equation}
Thus, ESAS is derived as follows,
\begin{equation}
\begin{split}
I_i(\mathcal{A};\mathcal{W}=w_i) = \frac{N_i}{N} \Bigl(1 + \frac{N_{L,i}}{N_i} \log(\frac{N_{L,i}}{N_i})+\frac{N_{H,i}}{N_i} \log(\frac{N_{H,i}}{N_i}) \Bigl)
\end{split}
\end{equation}

\subsection{Evaluating the effectiveness of ESAS metric}
\begin{figure}[t]
    \centering
    \begin{subfigure}{0.48\textwidth}
        \centering
        \includegraphics[width=\linewidth]{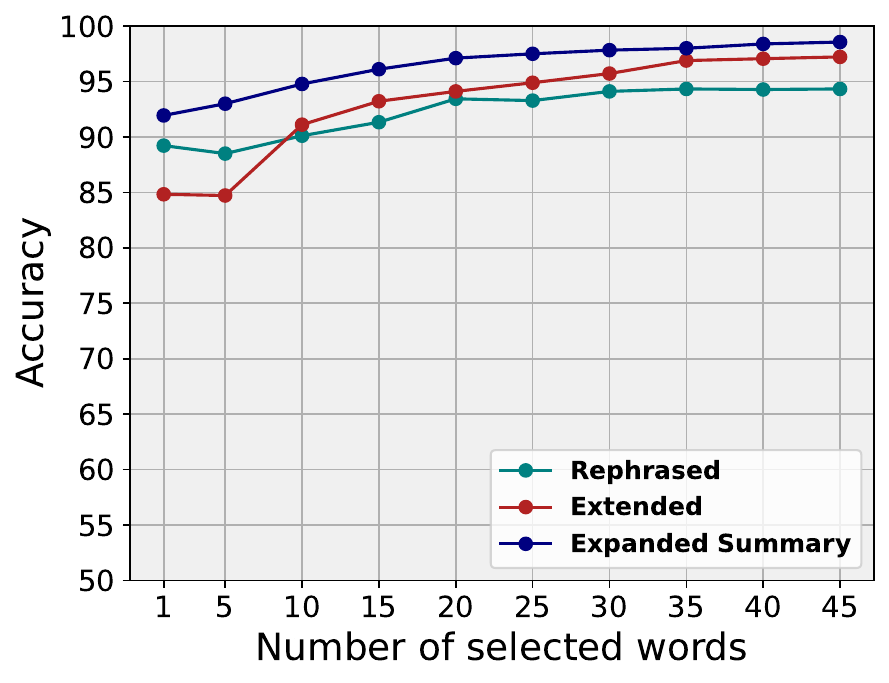}
        \caption{ChatGPT}
        \label{fig:plot1}
    \end{subfigure}
    \hfill
    \begin{subfigure}{0.48\textwidth}
        \centering
        \includegraphics[width=\linewidth]{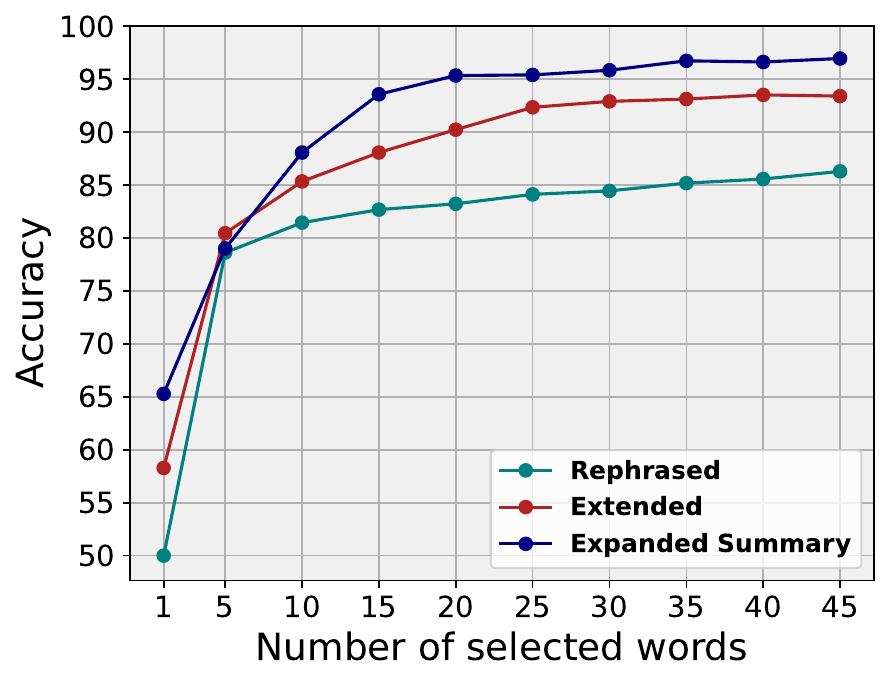}
        \caption{Llama2-7b}
        \label{fig:plot2}
    \end{subfigure}
    \medskip
    \begin{subfigure}{0.48\textwidth}
        \centering
        \includegraphics[width=\linewidth]{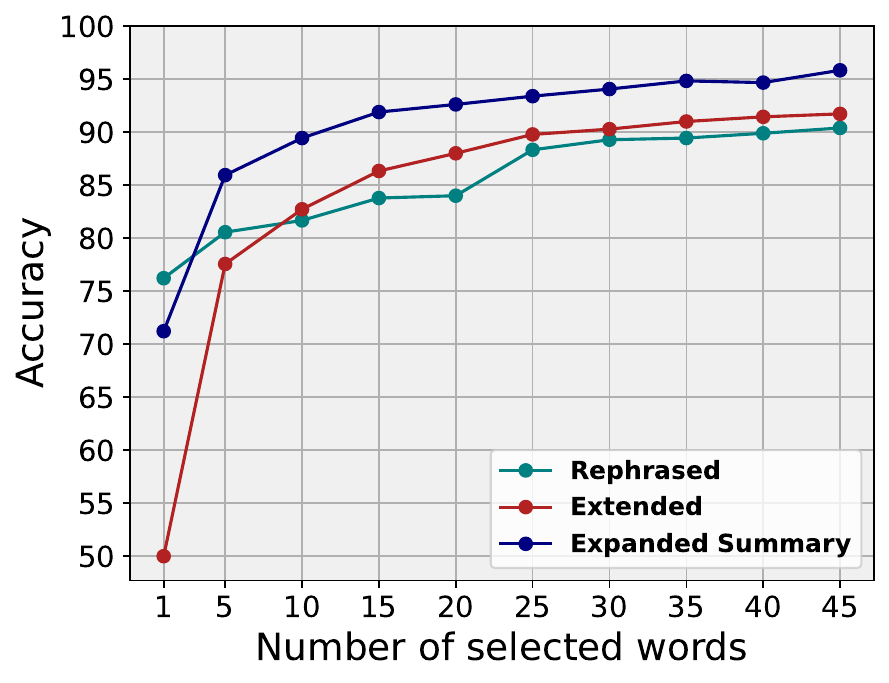}
        \caption{Llama2-13b}
        \label{fig:plot3}
    \end{subfigure}
    \hfill
    \begin{subfigure}{0.48\textwidth}
        \centering
        \includegraphics[width=\linewidth]{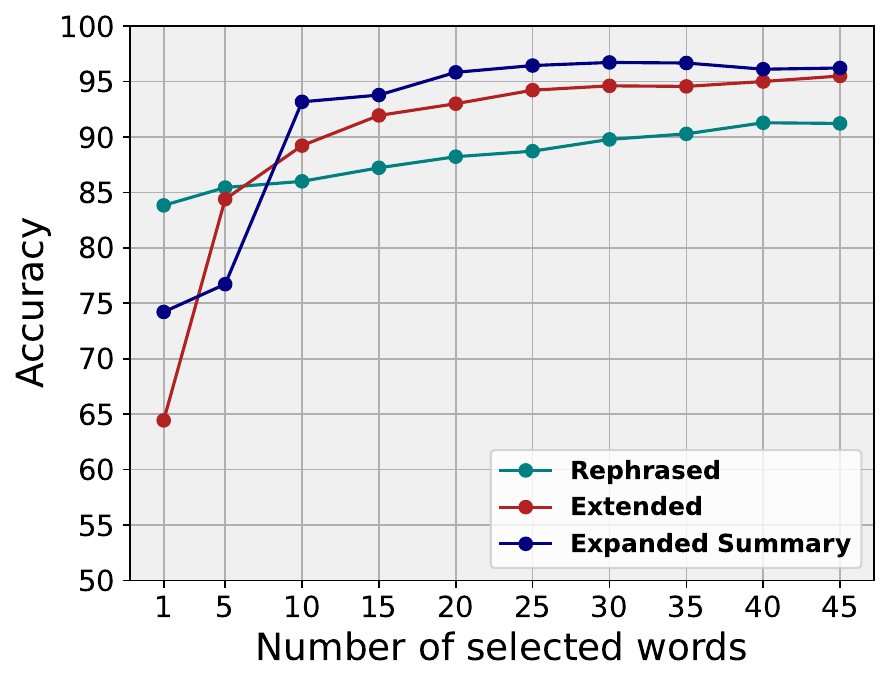}
        \caption{Mistral-7b}
        \label{fig:plot4}
    \end{subfigure}
    \caption{The accuracy of TF-IDF in classifying the authorship of news articles based on the number of words selected for its vocabulary using ESAS metric.}
    \label{fig:ESAS_eval}
\end{figure}
To evaluate the effectiveness of ESAS metric, we initially divide the data into a $70\%$ training set and a $30\%$ testing set.
Subsequently, we tokenize the news articles at the word level for both human-authored news articles (HANA) and LLM-generated news articles (LGNA).
The ESAS metric is then computed for each word within the combined vocabulary, followed by ranking them.
By selecting the $m$ words with the highest ESAS scores as the term vocabulary for the TF-IDF method, we train a binary classifier using logistic regression.
Fig. \ref{fig:ESAS_eval} displays the results of this trained classifier on the testing set's news articles across various LLMs and fake levels.
The results suggest that the classifier achieves high accuracy consistently when only a limited number of words are present in the TF-IDF vocabulary.
For $m{=}10$, an average accuracy of $92\%$, $84.9\%$, $84.6\%$, and $89.5\%$ are achieved for ChatGPT, Llama2-7b, Llama2-13b, and Mistral-7b, respectively.

Table \ref{tab:Voc_size} outlines the vocabulary sizes from the training data for every LLM and fake level.
The ``Common'' column indicates the number of words shared between the vocabularies of HANA and LGNA.
Conversely, the ``HANA uncommon'' and ``LGNA uncommon'' columns display the counts of words exclusive to the HANA and LGNA vocabularies, respectively.
Based on the data represented in the table, a size of $m{=}10$ constitutes merely $0.022\%$, $0.022\%$, $0.022\%$, and $0.021\%$ of the unique vocabulary sizes for ChatGPT, Llama2-7b, Llama2-13b, and Mistral-7b, respectively. Despite these minuscule percentages, the high accuracy achieved through the rudimentary classifier method confirm the ESAS metric's effectiveness in prioritizing and selecting significant words for discerning the authorship of news articles.

In addition, Fig. \ref{fig:ESAS_eval} illustrates varying levels of difficulty in predicting the labels for the three prompt strategies employed.
The ``Expanded summary'' strategy proves to be the easiest to detect as it achieves higher accuracy for a specific number of selected words, $m$.
Conversely, the ``Rephrased'' strategy emerges as the most challenging.
Across all LLMs, the ``Extended'' strategy curve lies between the other two.
This implies that our designed prompt strategies effectively generate distinct levels of fake.

\renewcommand{\arraystretch}{1}
\begin{table}[b]
\centering
\caption{The size of common and uncommon vocabulary for HANA and LGNA across different LLMs and fake levels}
\begin{tabular}{clccc}
\hline
\textbf{LLM} & \textbf{Prompt} & \textbf{Common}& \textbf{HANA uncommon}& \textbf{LGNA uncommon} \\
\hline
\multirow{3}{*}{\textbf{ChatGPT}} & Rephrased &34414 &9921 & 2168\\
                        & Extended &23692 &20643 & 2569\\
                        & Expanded summary &25874 &18461 & 2150\\
\hline
\multirow{3}{*}{\textbf{Llama2-7b}} & Rephrased & 27931 &16181 & 615\\
                        & Subrow 2 & 25097 & 19015& 2350\\
                        & Subrow 3 & 23282 & 20830& 1585\\
\hline
\multirow{3}{*}{\textbf{Llama2-13b}} & Rephrased & 28537& 15798& 683\\
                        & Extended & 26659 & 17676& 2522\\
                        &Expanded summary & 22404 & 21931& 1700\\
\hline
\multirow{3}{*}{\textbf{Mistral-7b}} & Rephrased & 33373 & 10962& 2481\\
                        & Extended &23412 & 20923& 4753\\
                        & Expanded summary  &26827  &17508 & 3708\\
\hline
\end{tabular}

\label{tab:Voc_size}
\end{table}

\section{Results and Discussion}
In this section, we extract cues from news articles based on unigrams, bigrams, and POS tagging across different LLMs in various scenarios.
These cues are derived from the 10 most significant entities (unigram, bigram, or POS tagging) selected using the ESAS metric from the entire training set of news articles.
We believe that by keeping these cues in mind while reading news articles on online social platforms, individuals can attain accuracy levels similar to the rudimentary classifier for identifying news articles which are likely generated by LLMs.

In all scenarios in this section, the data is divided into training and testing sets with a ratio of $70\%$ and $30\%$, respectively.

\renewcommand{\arraystretch}{1.1}
\begin{table}[b]
\centering
\caption{Top 10 words by ESAS metric. The left and right numbers in parentheses indicate frequency ratios to the most frequent word among the 10 selected for HANA and LGNA, respectively.}
\begin{tabular}{cll}
\hline
\textbf{LLM} & \textbf{Prompt} & \textbf{Words with highest ESAS value} \\
\hline
\multirow{6}{*}{\rotatebox[origin=c]{90}{\textbf{ChatGPT}}} & Rephrased & \textbf{said}(0.13,0.0), \textbf{it}(0.13,0.06), \textbf{we}(0.06,0.02), \textbf{is}(0.14,0.07), \textbf{was}(0.11,0.05), \\
&&\textbf{but}(0.06,0.02), \textbf{told}(0.02,0.0), \textbf{and}(0.42,0.31), \textbf{the}(1.0,0.83), \textbf{be}(0.07,0.03)\\\cline{2-3}

                        & Extended & \textbf{said}(1.0,0.04), \textbf{he}(0.76,0.25), \textbf{we}(0.49,0.14), \textbf{at}(0.66,0.24), \textbf{after}(0.26,0.05),\\
                        && \textbf{told}(0.14,0.0), \textbf{you}(0.29,0.06), \textbf{was}(0.83,0.4), \textbf{these}(0.09,0.31), \textbf{because}(0.14,0.01) \\\cline{2-3}
                        & Expanded summary & \textbf{said}(0.64,0.0), \textbf{we}(0.31,0.03), \textbf{he}(0.48,0.13), \textbf{you}(0.18,0.01), \textbf{it}(0.65,0.24),\\
                        && \textbf{was}(0.53,0.17), \textbf{that}(1.0,0.56), \textbf{told}(0.09,0.0), \textbf{so}(0.12,0.01), \textbf{because}(0.09,0.0)\\\cline{2-3}
\hline
\multirow{6}{*}{\rotatebox[origin=c]{90}{\textbf{Llama2-7b}}} & Rephrased & \textbf{the}(1.0,0.58), \textbf{said}(0.13,0.01), \textbf{of}(0.44,0.24), \textbf{to}(0.46,0.25), \textbf{in}(0.36,0.18), \\
&& \textbf{it}(0.13,0.04), \textbf{we}(0.06,0.01), \textbf{and}(0.42,0.26), \textbf{was}(0.11,0.04), \textbf{that}(0.2,0.11) \\\cline{2-3}
                        & Extended & \textbf{said}(0.23,0.12), \textbf{and}(0.77,1.0), \textbf{significant}(0.0,0.03), \textbf{at}(0.15,0.07), \textbf{you}(0.07,0.02), \\
                        && \textbf{has}(0.14,0.24), \textbf{conclusion}(0.0,0.02), \textbf{he}(0.18,0.1), \textbf{an}(0.11,0.06), \textbf{address}(0.01,0.03)\\\cline{2-3}
                        & Expanded summary & \textbf{said}(1.0,0.28), \textbf{he}(0.76,0.25), \textbf{you}(0.28,0.03), \textbf{was}(0.83,0.34), \textbf{significant}(0.02,0.18), \\
                        &&\textbf{at}(0.64,0.27), \textbf{conclusion}(0.0,0.13), \textbf{told}(0.14,0.01), \textbf{because}(0.14,0.01), \textbf{what}(0.23,0.05)\\\cline{2-3}
\hline
\multirow{6}{*}{\rotatebox[origin=c]{90}{\textbf{Llama2-13b}}} & Rephrased &\textbf{said}(0.13,0.02), \textbf{the}(1.0,0.63), \textbf{in}(0.36,0.17), \textbf{to}(0.46,0.26), \textbf{it}(0.13,0.04),\\
&&\textbf{of}(0.43,0.25), \textbf{we}(0.06,0.01), \textbf{he}(0.1,0.03), \textbf{was}(0.11,0.04), \textbf{on}(0.15,0.07)\\\cline{2-3}
                        & Extended & \textbf{and}(0.36,0.51), \textbf{has}(0.07,0.13), \textbf{said}(0.11,0.06), \textbf{significant}(0.0,0.02), \textbf{he}(0.08,0.04),\\
                        &&\textbf{you}(0.03,0.01), \textbf{the}(0.86,1.0), \textbf{at}(0.07,0.04), \textbf{however}(0.0,0.02), \textbf{com}(0.01,0.0) \\\cline{2-3}
                        &Expanded summary & \textbf{said}(0.36,0.07), \textbf{he}(0.27,0.07), \textbf{at}(0.23,0.08), \textbf{has}(0.21,0.42), \textbf{you}(0.1,0.02), \\
                        &&\textbf{it}(0.36,0.18), \textbf{was}(0.3,0.14), \textbf{told}(0.05,0.0), \textbf{in}(1.0,0.7), \textbf{significant}(0.01,0.06)\\\cline{2-3}
\hline
\multirow{6}{*}{\rotatebox[origin=c]{90}{\textbf{Mistral-7b}}} & Rephrased &\textbf{said}(0.13,0.01), \textbf{the}(1.0,0.67), \textbf{it}(0.13,0.04), \textbf{to}(0.46,0.29), \textbf{of}(0.43,0.27),\\
&&\textbf{in}(0.36,0.22), \textbf{that}(0.2,0.1), \textbf{is}(0.14,0.06), \textbf{he}(0.1,0.04), \textbf{we}(0.06,0.02)\\\cline{2-3}
                        & Extended &\textbf{said}(0.23,0.06), \textbf{at}(0.15,0.06), \textbf{he}(0.18,0.08), \textbf{you}(0.07,0.02), \textbf{and}(0.77,1.0), \\
                        && \textbf{was}(0.19,0.1), \textbf{because}(0.03,0.0), \textbf{told}(0.03,0.0), \textbf{after}(0.06,0.02), \textbf{year}(0.06,0.02) \\\cline{2-3}
                        & Expanded summary  & \textbf{said}(0.64,0.09), \textbf{he}(0.48,0.15), \textbf{you}(0.18,0.02), \textbf{it}(0.65,0.28), \textbf{was}(0.53,0.24),\\
                        && \textbf{because}(0.09,0.0), \textbf{told}(0.09,0.0), \textbf{that}(1.0,0.62), \textbf{so}(0.12,0.02), \textbf{significant}(0.01,0.1)\\\cline{2-3}
\hline
\end{tabular}

\label{tab:unigram}
\end{table}

\renewcommand{\arraystretch}{1}
\begin{table}[b]
\centering
\caption{Top 10 words by ESAS metric for different news topics. The left and right numbers in parentheses indicate frequency ratios to the most frequent word among the 10 selected for HANA and LGNA, respectively.}
\begin{tabular}{cll}
\hline
\textbf{LLM} & \textbf{Topic} & \textbf{Words with highest ESAS value} \\
\hline
\multirow{12}{*}{\rotatebox[origin=c]{90}{\textbf{ChatGPT}}} & History &  \textbf{said}(0.65,0.0), \textbf{he}(0.67,0.2), \textbf{we}(0.24,0.02), \textbf{was}(0.51,0.16), \textbf{that}(1.0,0.53),\\
&&\textbf{within}(0.03,0.2), \textbf{told}(0.09,0.0), \textbf{you}(0.1,0.0), \textbf{these}(0.05,0.23), \textbf{ongoing}(0.0,0.09)\\\cline{2-3}
                        & Politics &  \textbf{said}(1.0,0.0), \textbf{he}(0.97,0.21), \textbf{was}(0.85,0.2), \textbf{we}(0.37,0.02), \textbf{it}(0.84,0.29),\\
                        &&\textbf{concerns}(0.03,0.3), \textbf{significant}(0.02,0.26), \textbf{landscape}(0.01,0.22), \textbf{after}(0.39,0.07), \textbf{told}(0.16,0.0)\\\cline{2-3}
                        & Science & \textbf{said}(0.66,0.0), \textbf{we}(0.28,0.03), \textbf{it}(0.69,0.25), \textbf{that}(1.0,0.46), \textbf{you}(0.13,0.0),\\
                        && \textbf{was}(0.3,0.08), \textbf{so}(0.11,0.0), \textbf{potential}(0.04,0.2), \textbf{would}(0.17,0.03), \textbf{told}(0.08,0.0)\\\cline{2-3}
                        & Society &  \textbf{said}(1.0,0.01), \textbf{we}(0.51,0.05), \textbf{people}(0.5,0.07), \textbf{was}(0.84,0.26), \textbf{you}(0.25,0.0), \\
                        &&\textbf{he}(0.59,0.14), \textbf{told}(0.15,0.0), \textbf{individuals}(0.02,0.22), \textbf{because}(0.14,0.0), \textbf{it}(0.92,0.43)\\\cline{2-3}
                        & Sports & \textbf{said}(0.68,0.0), \textbf{we}(0.43,0.03), \textbf{you}(0.19,0.0), \textbf{it}(0.65,0.23), \textbf{was}(0.48,0.16), \\
                        &&\textbf{he}(0.37,0.1), \textbf{com}(0.15,0.01), \textbf{ap}(0.13,0.01), \textbf{that}(1.0,0.53), \textbf{because}(0.11,0.0)\\\cline{2-3}
\hline
\multirow{12}{*}{\rotatebox[origin=c]{90}{\textbf{Llama2-7b}}} & History &  \textbf{said}(0.98,0.27), \textbf{he}(1.0,0.32), \textbf{was}(0.77,0.31), \textbf{significant}(0.01,0.18), \textbf{ongoing}(0.0,0.13),\\
&&\textbf{towards}(0.01,0.16), \textbf{told}(0.14,0.01), \textbf{title}(0.01,0.13), \textbf{you}(0.15,0.01), \textbf{conclusion}(0.0,0.11)\\\cline{2-3}
                        & Politics & \textbf{said}(0.61,0.17), \textbf{he}(0.59,0.17), \textbf{was}(0.51,0.15), \textbf{significant}(0.01,0.15), \textbf{has}(0.57,1.0),\\
                        &&\textbf{would}(0.21,0.04), \textbf{told}(0.1,0.0), \textbf{concerns}(0.02,0.14), \textbf{had}(0.2,0.04), \textbf{were}(0.2,0.04)\\\cline{2-3}
                        & Science & \textbf{said}(1.0,0.21), \textbf{was}(0.45,0.11), \textbf{potential}(0.06,0.3), \textbf{he}(0.31,0.06), \textbf{conclusion}(0.0,0.14), \\
                        &&\textbf{you}(0.2,0.02), \textbf{significant}(0.02,0.21), \textbf{title}(0.0,0.12), \textbf{concerns}(0.07,0.29), \textbf{told}(0.12,0.0)\\\cline{2-3}
                        & Society & \textbf{said}(1.0,0.31), \textbf{he}(0.59,0.15), \textbf{was}(0.84,0.34), \textbf{you}(0.25,0.02), \textbf{title}(0.0,0.14),\\
                        &&\textbf{told}(0.15,0.01), \textbf{conclusion}(0.0,0.12), \textbf{had}(0.28,0.07), \textbf{she}(0.33,0.09), \textbf{towards}(0.02,0.16) \\\cline{2-3}
                        & Sports & \textbf{said}(1.0,0.32), \textbf{com}(0.22,0.0), \textbf{you}(0.28,0.02), \textbf{ap}(0.2,0.0), \textbf{was}(0.71,0.23), \\
                        &&\textbf{https}(0.18,0.0), \textbf{he}(0.55,0.18), \textbf{at}(0.7,0.28), \textbf{twitter}(0.15,0.0), \textbf{significant}(0.01,0.17)\\\cline{2-3}
\hline
\multirow{12}{*}{\rotatebox[origin=c]{90}{\textbf{Llama2-13b}}} & History &  \textbf{said}(0.37,0.07), \textbf{he}(0.38,0.1), \textbf{has}(0.19,0.39), \textbf{in}(1.0,0.67), \textbf{was}(0.29,0.13), \\
&&\textbf{at}(0.21,0.08), \textbf{significant}(0.01,0.07), \textbf{ongoing}(0.0,0.05), \textbf{told}(0.05,0.0), \textbf{what}(0.07,0.01)\\\cline{2-3}
                        & Politics &  \textbf{said}(0.55,0.1), \textbf{he}(0.53,0.13), \textbf{was}(0.46,0.15), \textbf{has}(0.51,1.0), \textbf{significant}(0.01,0.14),\\
                        &&\textbf{after}(0.22,0.04), \textbf{told}(0.09,0.0), \textbf{at}(0.29,0.09), \textbf{it}(0.46,0.21), \textbf{future}(0.03,0.15)\\\cline{2-3}
                        & Science & \textbf{said}(0.97,0.11), \textbf{he}(0.3,0.05), \textbf{was}(0.43,0.13), \textbf{at}(0.49,0.17), \textbf{significant}(0.02,0.2), \\
                        && \textbf{it}(1.0,0.53), \textbf{potential}(0.05,0.26), \textbf{what}(0.2,0.03), \textbf{told}(0.11,0.0), \textbf{has}(0.49,0.91)\\\cline{2-3}
                        & Society & \textbf{said}(0.34,0.08), \textbf{he}(0.2,0.05), \textbf{you}(0.09,0.0), \textbf{has}(0.18,0.39), \textbf{was}(0.29,0.14),\\
                        &&\textbf{told}(0.05,0.0), \textbf{in}(1.0,0.69), \textbf{what}(0.07,0.01), \textbf{people}(0.17,0.06), \textbf{at}(0.19,0.08) \\\cline{2-3}
                        & Sports & \textbf{said}(0.94,0.2), \textbf{com}(0.2,0.0), \textbf{https}(0.17,0.0), \textbf{at}(0.66,0.22), \textbf{ap}(0.19,0.0),\\
                        &&\textbf{he}(0.51,0.14), \textbf{has}(0.45,1.0), \textbf{you}(0.26,0.03), \textbf{was}(0.67,0.25), \textbf{because}(0.15,0.01)\\\cline{2-3}
\hline
\multirow{12}{*}{\rotatebox[origin=c]{90}{\textbf{Mistral-13b}}} & History & \textbf{said}(0.65,0.1), \textbf{he}(0.67,0.2), \textbf{was}(0.51,0.22), \textbf{title}(0.01,0.09), \textbf{told}(0.09,0.01), \\
&& \textbf{ongoing}(0.0,0.08), \textbf{that}(1.0,0.63), \textbf{because}(0.08,0.01), \textbf{you}(0.1,0.01), \textbf{significant}(0.01,0.09) \\\cline{2-3}
                        & Politics & \textbf{said}(1.0,0.14), \textbf{he}(0.97,0.26), \textbf{was}(0.85,0.34), \textbf{significant}(0.02,0.23), \textbf{title}(0.0,0.17), \\
                        &&\textbf{it}(0.84,0.37), \textbf{after}(0.39,0.1), \textbf{told}(0.16,0.01), \textbf{concerns}(0.03,0.21), \textbf{because}(0.12,0.0) \\\cline{2-3}
                        & Science & \textbf{said}(0.66,0.07), \textbf{it}(0.69,0.29), \textbf{potential}(0.04,0.22), \textbf{you}(0.13,0.01), \textbf{that}(1.0,0.56), \\
                        && \textbf{title}(0.0,0.1), \textbf{significant}(0.02,0.13), \textbf{he}(0.21,0.05), \textbf{told}(0.08,0.0), \textbf{so}(0.11,0.01)\\\cline{2-3}
                        & Society &  \textbf{said}(1.0,0.14), \textbf{he}(0.59,0.17), \textbf{title}(0.0,0.17), \textbf{you}(0.25,0.03), \textbf{was}(0.84,0.36), \\
                        && \textbf{people}(0.5,0.17), \textbf{told}(0.15,0.01), \textbf{it}(0.92,0.46), \textbf{because}(0.14,0.01), \textbf{they}(0.51,0.21)\\\cline{2-3}
                        & Sports & \textbf{said}(1.0,0.13), \textbf{you}(0.28,0.02), \textbf{it}(0.96,0.4), \textbf{he}(0.55,0.15), \textbf{com}(0.22,0.01), \\
                        && \textbf{ap}(0.2,0.01), \textbf{was}(0.71,0.27), \textbf{https}(0.18,0.01), \textbf{because}(0.16,0.0), \textbf{that}(1.0,0.57)\\\cline{2-3}
\hline
\end{tabular}
\label{tab:topic}
\end{table}

\subsection{Cues based on 10 most significant unigrams for different level of fake} \label{sec:5.1}
We tokenize the news articles in the training set at the unigram level and apply the ESAS metric to all unique unigrams in the aggregated vocabularies of HANA and LGNA.
The unigrams are sorted based on their ESAS scores, and the top 10 words with the highest scores are selected.
Table \ref{tab:unigram} presents selected words in descending order with respect to ESAS score for each LLM and prompt strategy.
For each word among these 10 selected words, we calculated its frequency ratio in HANA or LGNA relative to the word with the highest frequency among the words selected.
The first number in parentheses represents this ratio for HANA, while the second number indicates the ratio for LGNA.

The table reveals that in 10 out of 12 instances, the first chosen word is \textit{``said''}. 
The frequency of \textit{``said''} in HANA is significantly higher than its frequency in LGNA.
This indicates that the LLMs under examination tend to use the word \textit{``said''} less frequently in their generated news articles, while human authors are more inclined to use it more often in reporting news.
Comparing the results with the curves shown in Fig. \ref{fig:ESAS_eval} reveals that the presence or absence of the word \textit{``said''} in a news article ($m{=}1$) can achieve an accuracy of over $85\%$ for news generated by ChatGPT across all prompt strategies.
The ratios of $0$, $0.02$, and $0$ for the three strategies in ChatGPT's results for the word \textit{``said''} confirm that ChatGPT is unlikely to use this word in its news articles.
Similarly, the word \textit{``told''} is more prevalent in HANA, featuring in 7 out of 12 instances, while its relative frequency ratio in LGNA is nearly zero.
Additionally, the frequency of most selected words is greater in HANA compared to LGNA. However, there are some terms with notably higher frequencies in LGNA that can assist individuals in recognizing articles produced by LLMs, like \textit{``significant''}, \textit{``address''}, \textit{``conclusion''}, and \textit{``however''}.

A noteworthy observation from the table is the abundance of stop words selected as the most significant words.
While crafting straightforward rules for news readers based on stop words is challenging, their role in providing discriminative features for such detection should not be overlooked.
Therefore, detection systems that exclude stop words in preprocessing stages should assess this impact on their performance.

\begin{figure}[t]
    \centering
    \begin{subfigure}{0.48\textwidth}
        \centering
        \includegraphics[width=\linewidth]{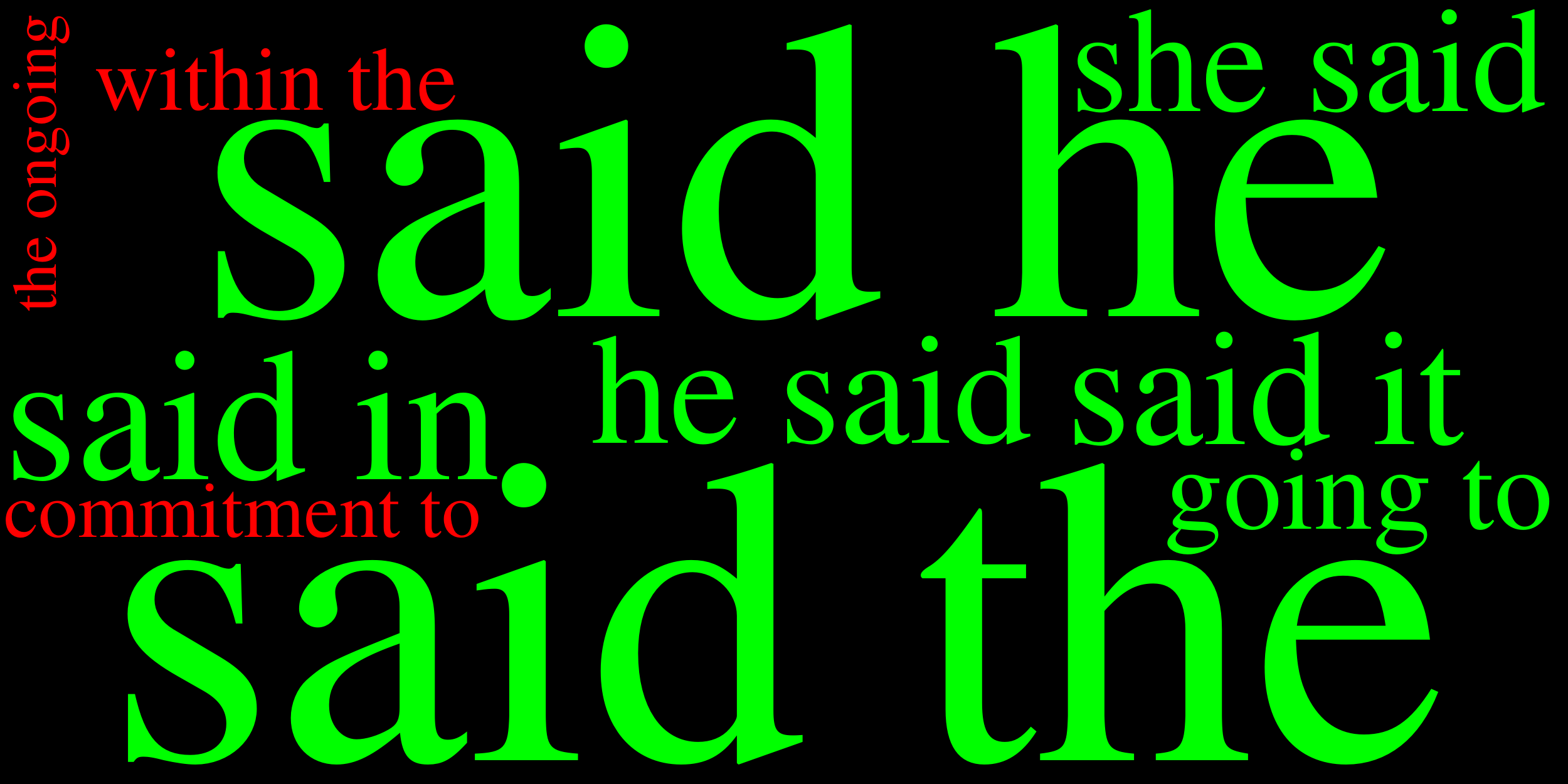}
        \caption{ChatGPT}
        \label{fig:plot1}
    \end{subfigure}
    \hfill
    \begin{subfigure}{0.48\textwidth}
        \centering
        \includegraphics[width=\linewidth]{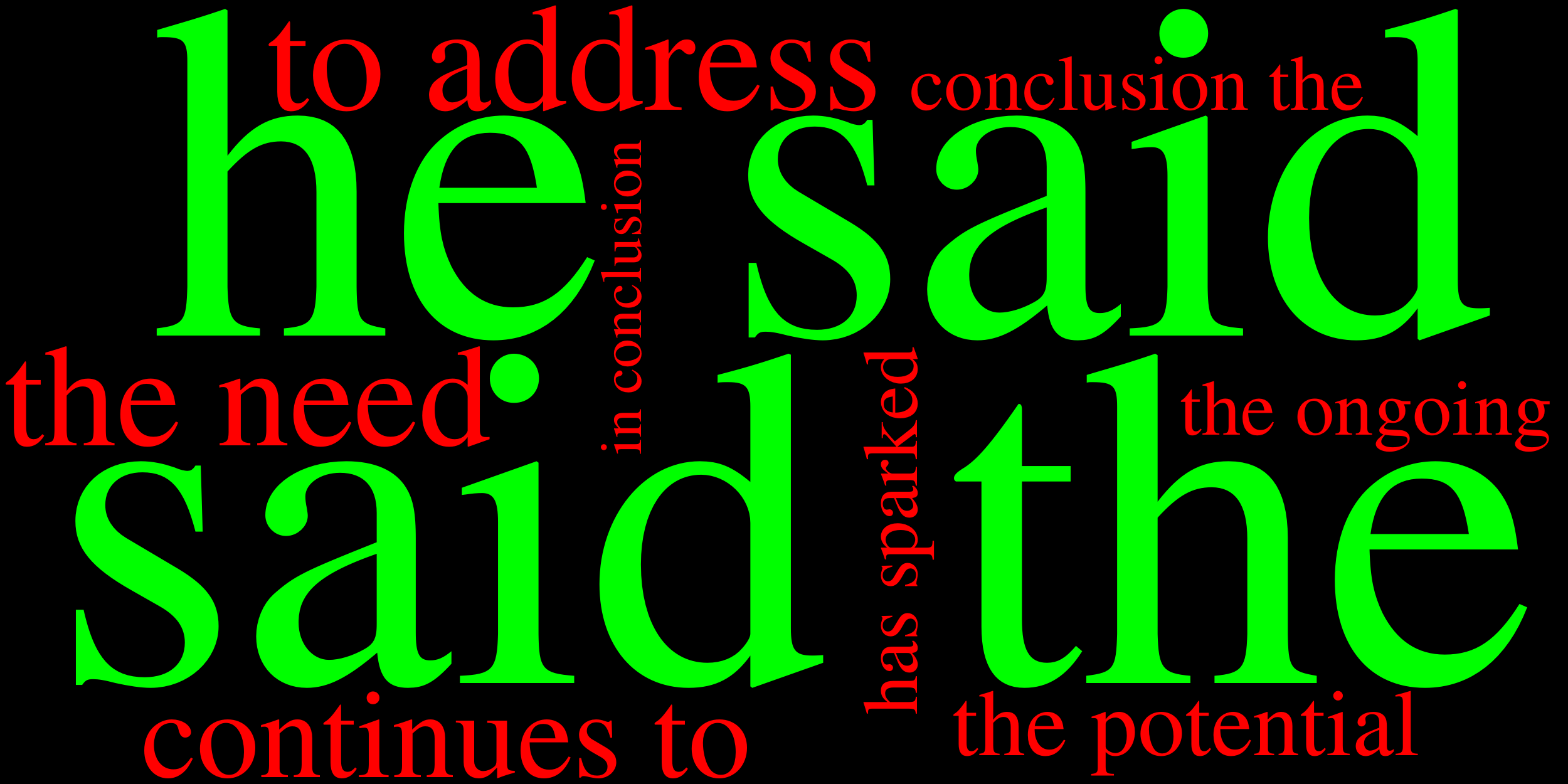}
        \caption{Llama2-7b}
        \label{fig:plot2}
    \end{subfigure}
    \medskip
    \begin{subfigure}{0.48\textwidth}
        \centering
        \includegraphics[width=\linewidth]{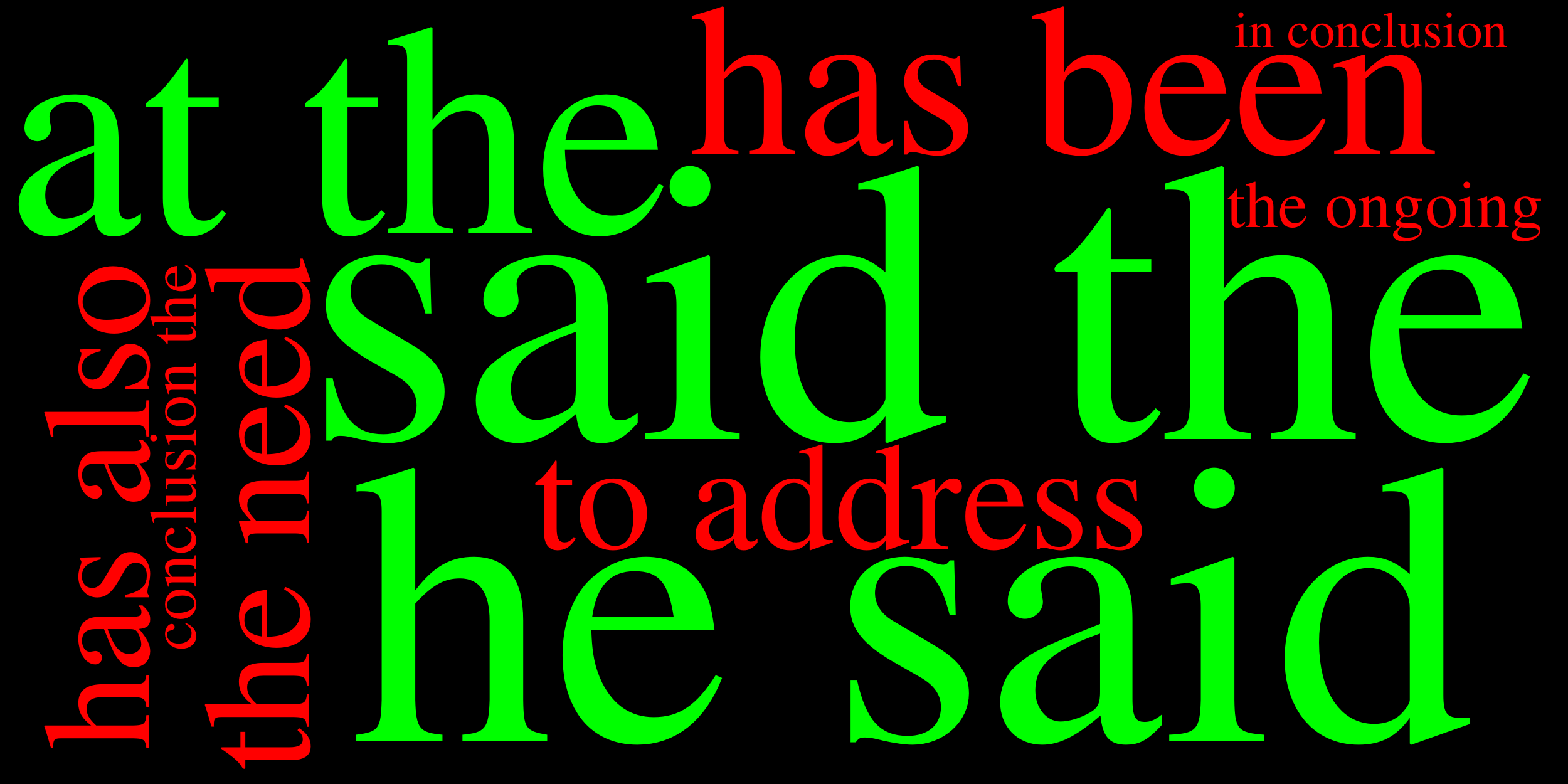}
        \caption{Llama2-13b}
        \label{fig:plot3}
    \end{subfigure}
    \hfill
    \begin{subfigure}{0.48\textwidth}
        \centering
        \includegraphics[width=\linewidth]{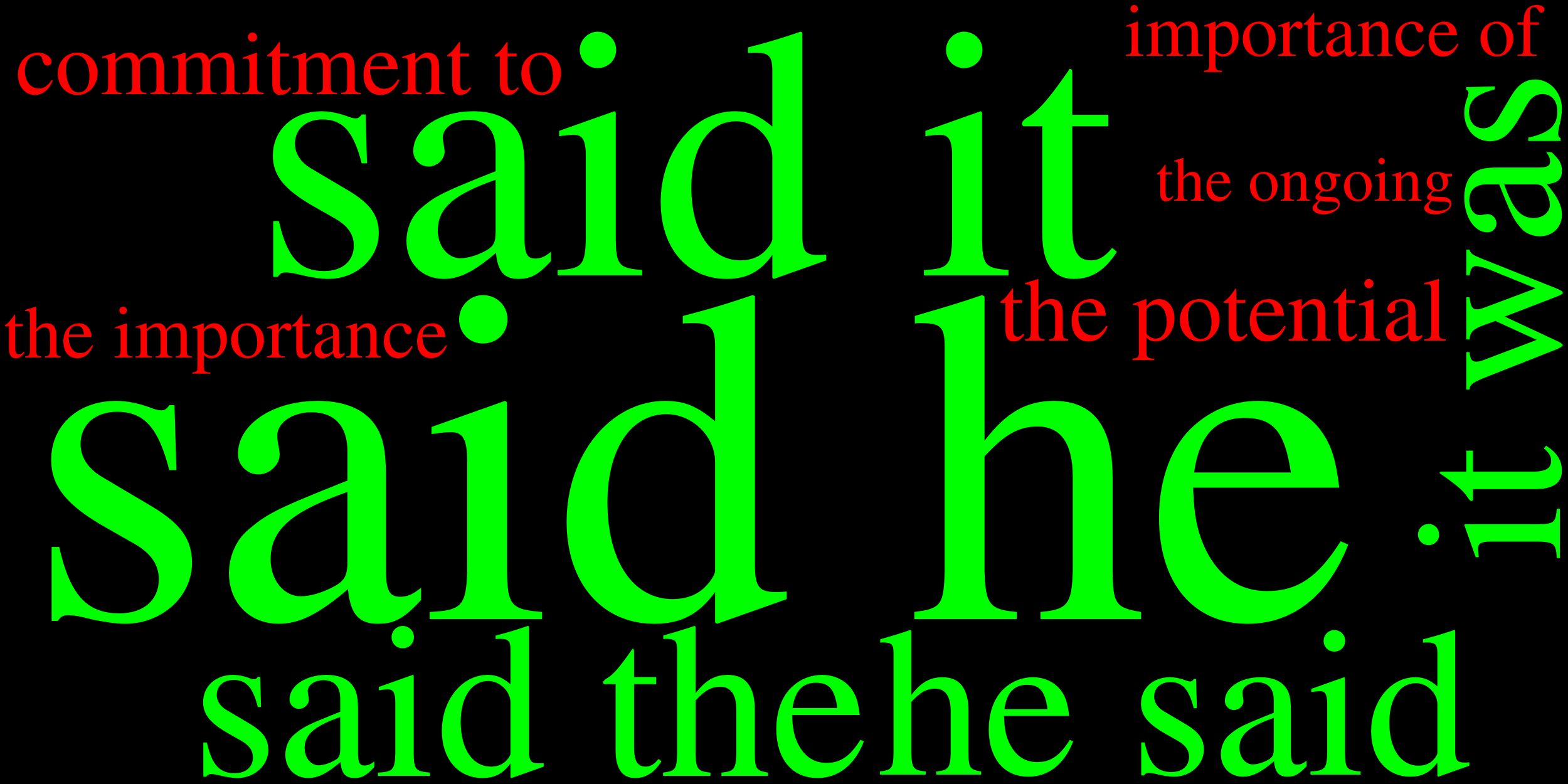}
        \caption{Mistral-7b}
        \label{fig:plot4}
    \end{subfigure}
    \caption{The word cloud of 10 most significant bigrams, with font size representing the relative ESAS score. Terms highlighted in green and red indicate higher relative frequencies in HANA and LGNA, respectively.}
    \label{fig:cloud}
\end{figure}

\begin{figure}[t]
    \centering
    \begin{subfigure}{0.48\textwidth}
        \centering
        \includegraphics[width=\linewidth]{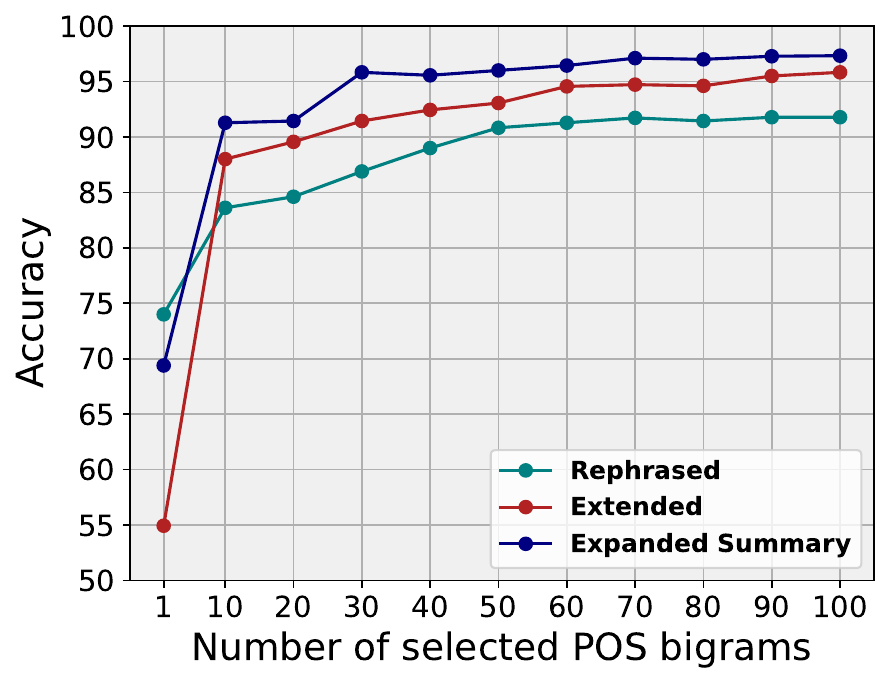}
        \caption{ChatGPT}
        \label{fig:plot1}
    \end{subfigure}
    \hfill
    \begin{subfigure}{0.48\textwidth}
        \centering
        \includegraphics[width=\linewidth]{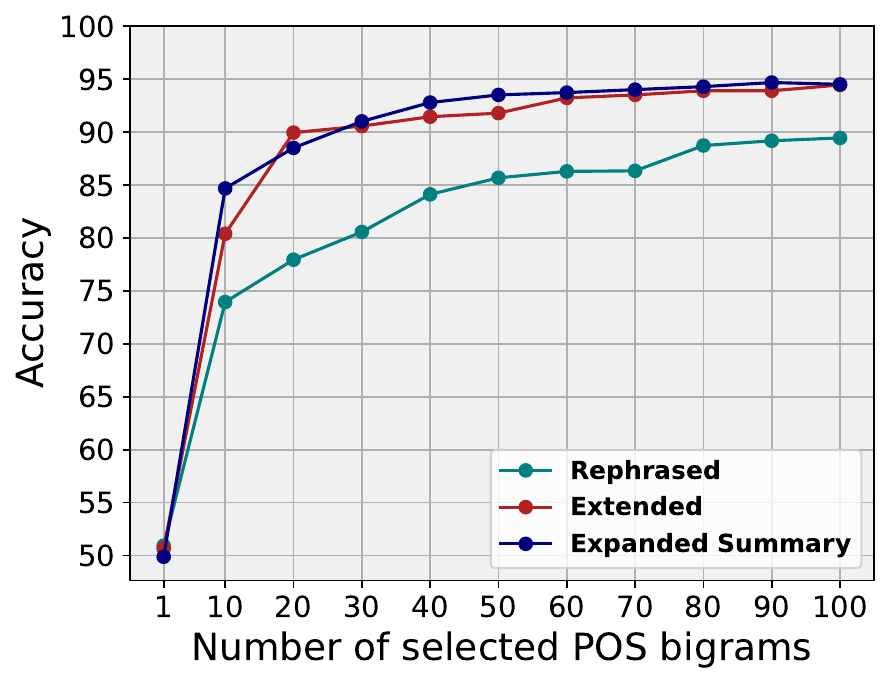}
        \caption{Llama2-7b}
        \label{fig:plot2}
    \end{subfigure}
    \medskip
    \begin{subfigure}{0.48\textwidth}
        \centering
        \includegraphics[width=\linewidth]{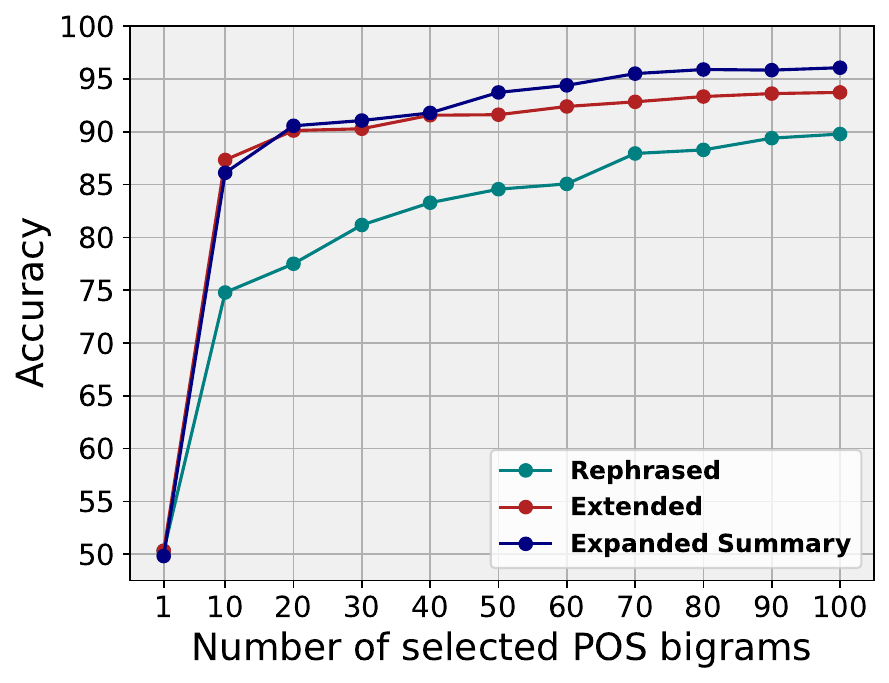}
        \caption{Llama2-13b}
        \label{fig:plot3}
    \end{subfigure}
    \hfill
    \begin{subfigure}{0.48\textwidth}
        \centering
        \includegraphics[width=\linewidth]{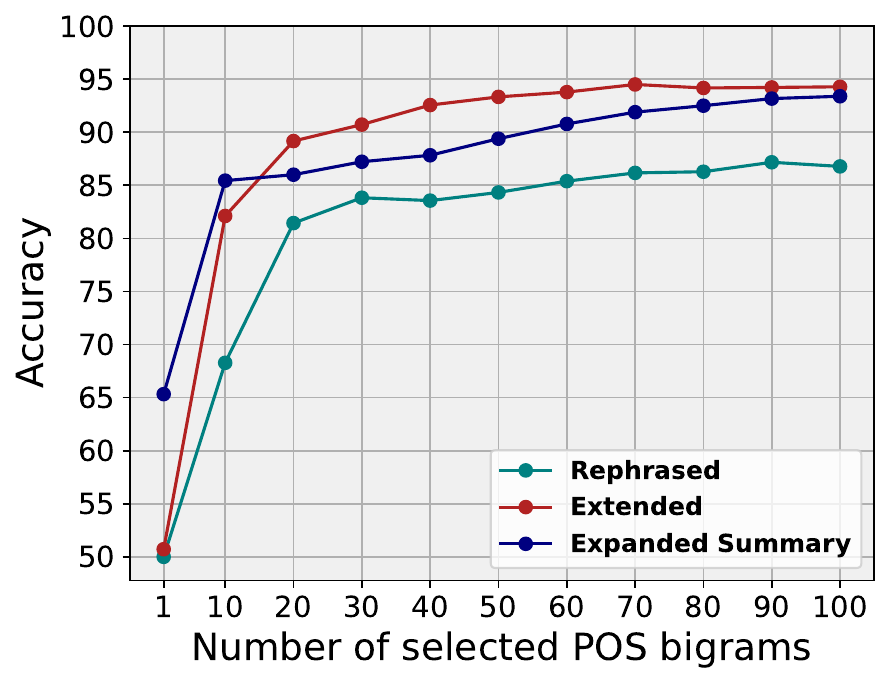}
        \caption{Mistral-7b}
        \label{fig:plot4}
    \end{subfigure}
    \caption{The accuracy of TF-IDF in classifying the authorship of news articles based on the number of POS bigrams selected for its vocabulary using ESAS metric.}
    \label{fig:posacc}
\end{figure}

\subsection{Cues based on 10 most significant unigrams for different news topics}\label{sec:5.2}
Research has shown that the performance of AI detection systems deteriorates when the field or topic of the text changes.
In this section, we investigate the impact of concentrating on a specific news topic on the selected unigrams for distinguishing between HANAs and LGNAs.
To achieve this, at each step, we filter one topic and construct a training set comprising only that particular topic.
Following this, we tokenize the news articles in the training set at the unigram level.
The ESAS metric is applied to the aggregated vocabulary, and the unigrams are sorted according to their ESAS scores. We subsequently choose the 10 most significant unigrams.
Due to limited space, we focus solely on the LGNA that are produced from the \textit{``Expanded Summary''} strategy in this part.
We opted for the \textit{``Expanded Summary''} strategy over the other two strategies as we attained higher accuracy with TF-IDF using this particular prompt strategy. 
Hence, we have greater confidence in the terms selected using the ESAS metric in this scenario.
Table \ref{tab:topic} presents the top 10 words selected for each news topic across each LLM.

In all cases, the word \textit{``said''} achieves the highest ESAS score. 
Among the LLMs, Llama2-7b displays a stronger inclination towards using the word \textit{``said''}, particularly in sports news.
However, its frequency is still approximately three to four times less frequent than in HANA.
Consequently, the presence of the word \textit{``said''} remains a pivotal indicator that the news article is human-written.
Similarly, as in Section \ref{sec:5.1}, the words \textit{``told''} and \textit{``significant''} offer valuable insights into the news source regardless of the topic. 
For the former, the HANA frequency surpasses that of LGNA, while for the latter, the frequency in LGNA exceeds that of HANA.

From the table, it is evident that particular words emerge when focusing on individual topics.
For instance, in sports-related news articles, tokens like \textit{``com''} and \textit{``https''} can achieve high ESAS scores.
Their HANA frequency ratio significantly surpasses their LGNA frequency ratio.
This suggests that the presence of a website link may indicate that the news article is authored by a human.
In history-related news, the word \textit{``ongoing''} is found in LGNA about five times more often than in HANA. In the context of science news, LLMs favor the use of \textit{``potential''} over human authors, with a frequency approximately five times higher. Meanwhile, human authors use the word \textit{``people''} more frequently in science articles.
In political news, new terms do not show consistent patterns.
For ChatGPT, Llama2-13b, and both Llama2-7b and Mistral-7b, the terms \textit{``landscape''}, \textit{``future''} and \textit{``concern''} appear, respectively.

\subsection{Cues based on 10 most significant bigrams}
We tokenize the news articles in the training set at the bigram level and apply the ESAS metric to all unique bigrams in the aggregated vocabularies of HANA and LGNA.
We sort the bigrams based on their ESAS scores and select the top 10 with the highest scores.
Similar to Section \ref{sec:5.2}, we focus on the \textit{``Expanded Summary''} strategy.
Fig. \ref{fig:cloud} presents the word cloud of selected bigrams for each LLM, with font size indicating their relative ESAS scores. Bigrams more frequent in HANA are shown in green, while those more frequent in LGNA are displayed in red.

Across all LLMs, bigrams containing the word \textit{``said"''} make the most significant contribution to identifying the authorship of a news article.
Specifically, bigrams like \textit{``he said''}, \textit{``said he''}, \textit{``said it''}, \textit{``said the''}, \textit{``she said''}, and \textit{``said in''} are common in human-authored news writing. 
The only bigram consistently present across all LLMs for discerning LLM-generated news is \textit{``the ongoing''}.
Additionally, each LLM presents its unique key bigram cues for identification as LLM-generated.
ChatGPT frequently employs the bigram \textit{``within the''}, and \textit{``commitment to''} which is also found in Mistral-7b. 
Additionally, Mistral-7b uses bigrams featuring the word \textit{``importance''}, including \textit{``the importance''} and \textit{``importance of''}.
While Llama2-7b and Llama2-13b share several common bigrams, such as \textit{``to address''}, \textit{``the need''} and \textit{``conclusion the''} they also reveal unique bigrams not shared between them.
In LGNA produced by Llama2-7b, bigrams like \textit{``continues to''} and \textit{``has sparked''} serve as strong indicators for identifying the news source as an LLM. Conversely, in LGNA generated by Llama2-13b, bigrams such as \textit{``has been''}, and \textit{``has also''} are crucial for distinguishing LLM-generated articles.

The findings demonstrate that the complexity of developing a detection method is influenced by both the model and its parameter count. Consequently, to construct a more effective detection system, training should encompass diverse samples from multiple LLMs, and samples from each LLM should be drawn from various settings.
\begin{figure}[t]
    \centering
    \begin{subfigure}{0.48\textwidth}
        \centering
        \includegraphics[width=\linewidth]{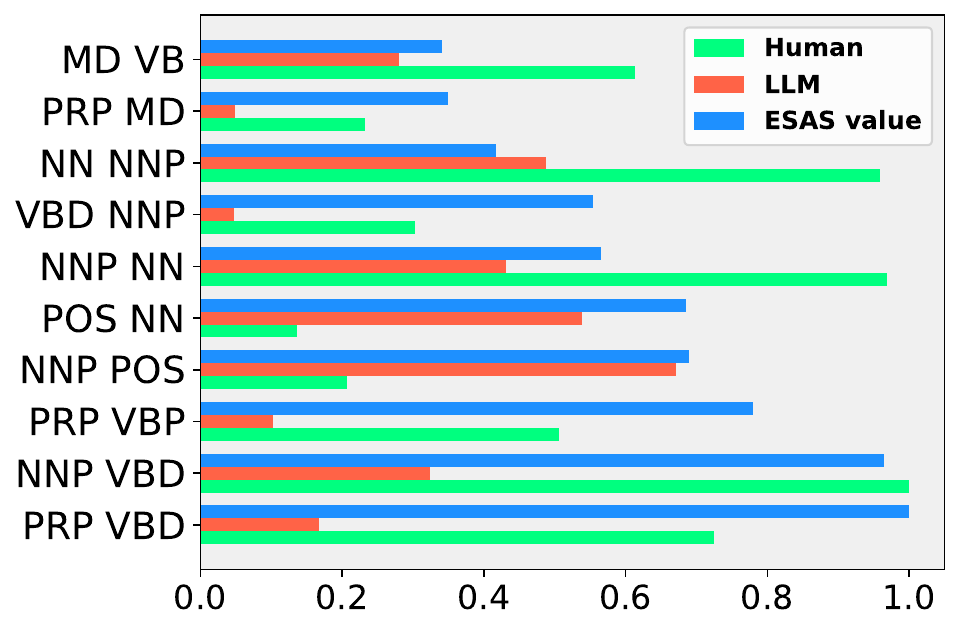}
        \caption{ChatGPT}
        \label{fig:plot1}
    \end{subfigure}
    \hfill
    \begin{subfigure}{0.48\textwidth}
        \centering
        \includegraphics[width=\linewidth]{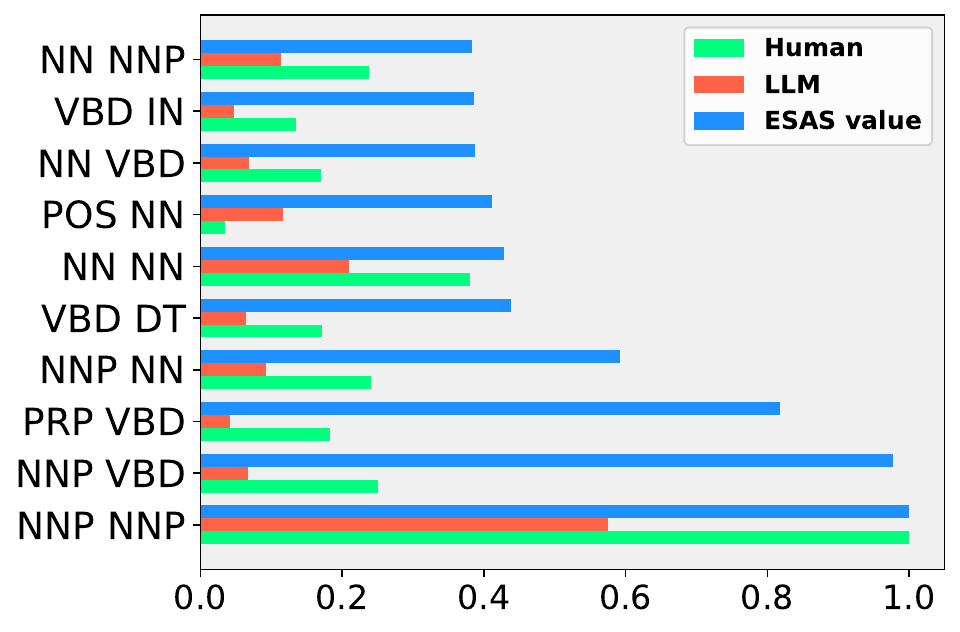}
        \caption{Llama2-7b}
        \label{fig:plot2}
    \end{subfigure}
    \medskip
    \begin{subfigure}{0.48\textwidth}
        \centering
        \includegraphics[width=\linewidth]{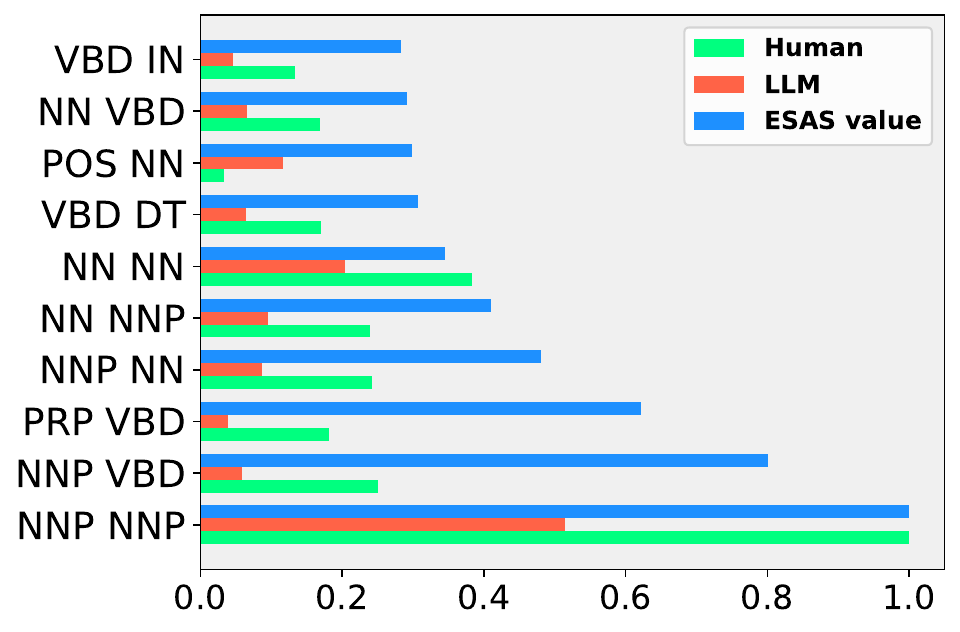}
        \caption{Llama2-13b}
        \label{fig:plot3}
    \end{subfigure}
    \hfill
    \begin{subfigure}{0.48\textwidth}
        \centering
        \includegraphics[width=\linewidth]{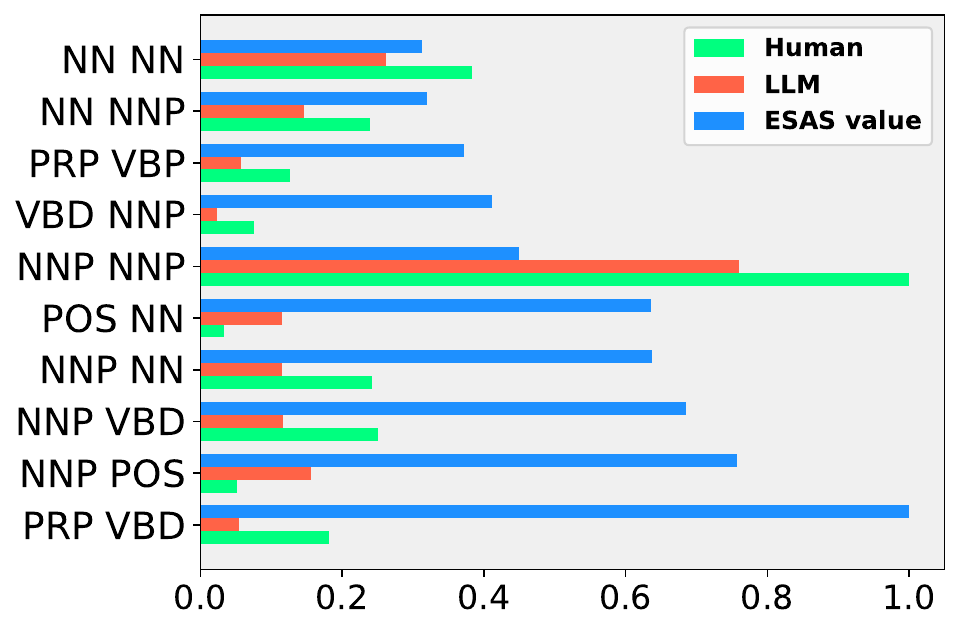}
        \caption{Mistral-7b}
        \label{fig:plot4}
    \end{subfigure}
    \caption{The barplots of 10 most significant POS bigrams across different LLMs for ``Extended summary'' prompt strategy.}
    \label{fig:barplots}
\end{figure}
\subsection{Cues based on 10 most significant bigrams in POS tagging}
Initially, we extract the POS tagging of sentences in each news article from our training set using Natural Language Toolkit (NLTK) POS tagging \cite{loper2002nltk} in Python. Subsequently, we employ a sliding window that combines two consecutive tags into a single entity, referred to as POS bigrams. We apply the ESAS metric to all unique POS bigrams in the aggregated vocabularies of HANA and LGNA. The POS bigrams are then sorted based on their ESAS scores, and the top 10 POS bigrams with the highest scores are selected.
Fig. \ref{fig:posacc} shows the accuracy of the TF-IDF combined with logistic regression classifier on the testing set's news articles across different LLMs and levels of fake. 
While the classifier attains reasonably high accuracy, it is lower than the accuracy explored in Section \ref{sec:5.1}.

In Fig. \ref{fig:barplots}, the 10 most significant POS bigrams for the "Extended summary" prompt strategy are shown across different LLMs.
The definitions for each tag are provided in Table \ref{tab:posset}.
For almost all POS bigrams, human writers use each entity more often than LLM-generated articles.
The exception arises when the \textit{``POS''} tag, representing a genitive marker, appears in the POS bigrams.
All LLMs tend to use \textit{``POS NN''} more frequently than human writers.
The ChatGPT and Mistral-7b models also utilize the POS bigram \textit{``NNP POS''} alongside \textit{``POS NN''}.
The genitive case usually denotes ownership or possession and is represented by an apostrophe followed by an "s" or simply an apostrophe (').
Hence, identifying the apostrophe used for possession can serve as a useful cue to be more skeptical about the possibility of the news article being generated by an LLM.
\begin{table}[t]
    \centering
     \caption{POS tag set}
    \begin{tabular}{cc|cc}
    \hline
      \textbf{POS Tag}   &  \textbf{Definition}&\textbf{POS Tag}   &  \textbf{Definition}\\\hline
     DT&determiner&POS& genitive marker\\
IN& preposition or conjunction, subordinating&PRP& pronoun, personal\\
MD& modal auxiliary&VB& verb, base form\\
NN& noun, common, singular or mass&VBD& verb, past tense\\
NNP& noun, proper, singular&VBP& verb, present tense, not 3rd person singular\\\hline
    \end{tabular}
    \label{tab:posset}
\end{table}

\section{Conclusion and Future Work}
In this study, we introduced the Entropy-Shift Authorship Signature (ESAS), a metric designed to rank terms and entities, such as POS tagging, within news articles based on their importance in distinguishing between human-written and LLM-generated news.
Furthermore, we presented our collected news dataset comprising 39k news articles authored by humans or produced using four LLMs, i.e. ChatGPT, Llama2-7b, Llama2-13b, and Mistral-7b, across three levels of fake.
We showcased the effectiveness of the proposed ESAS metric in in identifying significant indicators within the news articles. 
This was demonstrated by the impressive accuracy attained by a basic method, namely TF-IDF combined with a logistic regression classifier, when provided with a limited set of terms with the highest ESAS scores.
We analyzed and presented these top-ranked ESAS terms as straightforward cues that individuals can utilize to increase their skepticism towards LLM-generated fake news.
One significant area yet to be explored in our research is the situations where imposters utilize LLMs to create fake news and then manipulate them before publishing. This poses a question about the potential consequences of manipulating LLM-generated fake news. 
We defer this matter to future studies.
\begin{acks}
Research was supported in part by grant ARO W911NF-20-1-0254. The views and conclusions contained in this document are those of the authors and not of the sponsors. The U.S. Government is authorized to reproduce and distribute reprints for Government purposes notwithstanding any copyright notation herein.
\end{acks}

\bibliographystyle{ACM-Reference-Format}
\bibliography{paper}
\end{document}